# Is there a half-life for the success rates of AI agents?

Toby Ord
Oxford Martin AI Governance Initiative
University of Oxford

Building on the recent empirical work of Kwa et al. (2025), I show that within their suite of research-engineering tasks the performance of AI agents on longer-duration tasks can be explained by an extremely simple mathematical model — a constant rate of failing during each minute a human would take to do the task. This implies an exponentially declining success rate with the length of the task and that each agent could be characterised by its own half-life. This empirical regularity allows us to estimate the success rate for an agent at different task lengths. And the fact that this model is a good fit for the data is suggestive of the underlying causes of failure on longer tasks — that they involve increasingly large sets of subtasks where failing any one fails the task. Whether this model applies more generally on other suites of tasks is unknown and an important subject for further work.

## METR's results on the length of tasks agents can reliably complete

A recent paper by [Kwa et al. (2025)](#) from the research organisation METR has found an exponential trend in the duration of the tasks that frontier AI agents can solve: every 7 months, the length of task they can solve doubles.

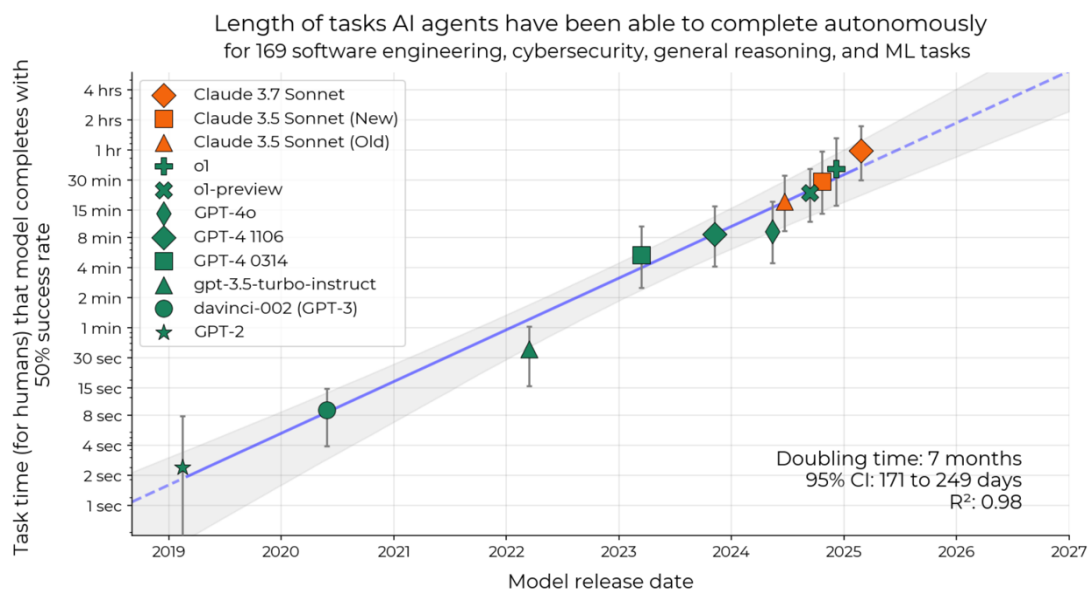

*Figure 1.* The exponential increase in achievable task duration over time as frontier models improve (reproduced from Kwa et al.).

These headline results are based on a test suite of 170 software engineering, cybersecurity, general reasoning, and ML tasks that they assembled to be indicative of the kinds of tasks that could help AI agents assist in AI research. These tasks are assembled from three different benchmarks that take different amounts of time for humans to achieve:

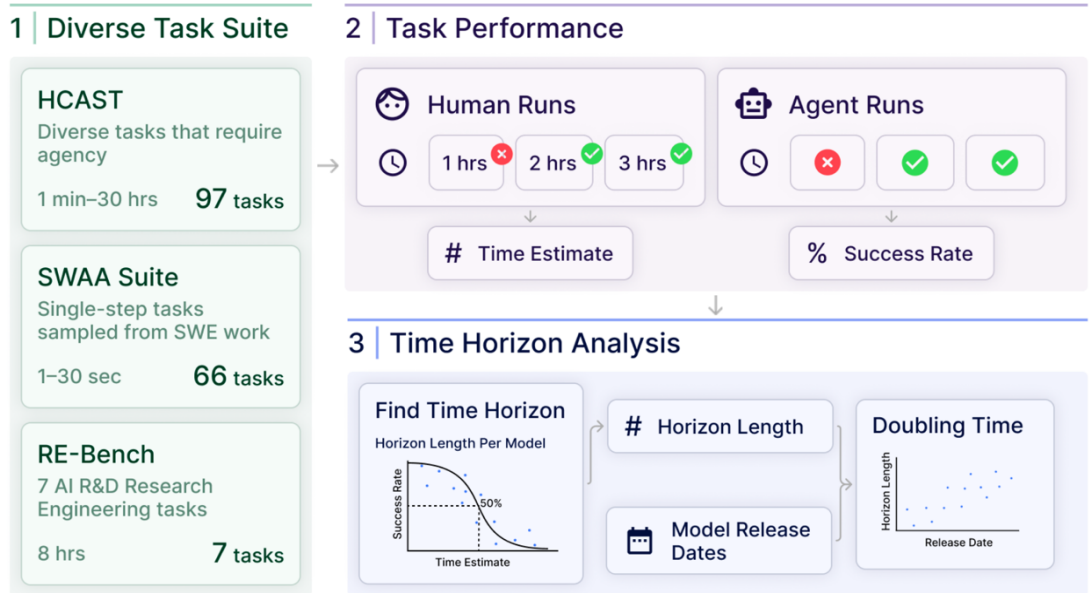

*Figure 2.* The task suite and analysis outline of Kwa et al.

In general, ability to perform a task drops off as its duration increases, so they use the AI agent's performance on tasks of different lengths to estimate the task-length at which the model would have a 50% success rate. They then showed that this length has been doubling every 7 months as the capabilities of frontier agents improve. The task-lengths are measured by how long it took humans to solve the same tasks.

They used 50% success rate as their chief performance threshold because it is the easiest level to robustly estimate. They are well aware that for many tasks, the required success rate for useful work may be much higher — such as 80%, 99%, or 99.9999%. They do measure the 80% success rate and find their mean estimate to have a doubling time of 213 days, compared to 212 days of the 50% rate. These are close to identical within their margin of error (±40 days for the 50% rate), so they conclude that the particular threshold doesn't seem to have much effect on their headline result about the rate of improvement.

But there is quite a gap between the 50% success rate time-horizon and the 80% success rate time horizon. For the best model (Claude 3.7 Sonnet) it could achieve a 50% success rate on tasks up to 59 minutes *vs* only 15 minutes if an 80% success rate was required. If those results generalise to the other models, then we could also see it like this: the task length for an 80% success rate is 1/4 the task length for a 50% success rate. Or in terms of improvement: what is doable with a 50% success rate now is doable with an 80% success rate in 14 months' time (= 2 doubling times).

The idea of measuring improvement in AI capabilities over time via time horizons at a chosen success rate is novel and interesting. AI forecasting is often hamstrung by the lack of a good measure for the y-axis of performance over time. We can track progress within a particular benchmark, but these are often solved in a couple of years, and we lack a good measure of underlying capability that can span multiple benchmarks. METR's measure allows comparisons between very different kinds of tasks in a common currency (time it takes a human) and shows a strikingly clear trend line — suggesting it is measuring something real.

But there are grounds for criticism too. In particular, there is room to wonder how much these results generalise outside of this kind of task suite. We know that there are some tasks humans can do very quickly that AIs can't solve (e.g. some simple spatial reasoning or intuitive physics tasks) and others that would take humans an extremely long time, but AIs can do quickly (e.g. rote mathematics). So a simple measure of 'time it would take a human' cannot explain all AI capability improvements. Kwa et al. are aware of this and even list several other ways that this task-suite may be non-representative of real-world performance including:

- All tasks were automatically scorable (a domain where RL works best)
- No interaction with other agents
- Lax resource constraints

For the purposes of this essay, we will take the data for what it is (performance on a particular task suite that may or may not generalise further) and explore underlying mechanisms that could explain it.

## Explaining these results via a constant hazard rate

These results call out for some explanation of what is going on. For example, exactly how does the time horizon shrink as the required success probability is increased? And what does it *mean* for an agent to be able to perform an 8-hour task, but not a 16-hour task. Isn't a 16-hour task just one 8-hour task after another?

Survival analysis is the field of understanding how the probability of something failing increases as a function of time. It tracks the survival probability at a time $S(t)$ — that is, the chance it still hasn't failed by that point. The simplest model in survival analysis is a constant hazard rate. This means that the chance of something failing in the next step (conditional on making it that far) is constant. A constant hazard rate leads to an exponentially declining survival curve. This behaviour is well-known from phenomena like radioactive decay, where there is a constant chance of decay at any moment, leading to an exponentially declining chance of the isotope's survival over time, which is often measured by a half-life.

If AI agent success-rates drop off with task length in this manner, then the 50% success rate time-horizon for each agent from Kwa et al. is precisely the *half-life* of that agent. As with the half-life of a radioisotope, this isn't just the median lifespan, it is the median remaining lifespan starting at any time — something that is only possible for an exponential survival curve. Unlike for particles, this AI agent half-life would be measured not in clock time, but in how long it takes a human to complete the task.

One rationale for this constant hazard rate model for AI agents is that tasks require getting past a series of steps each of which could end your attempt, with the longer the duration of the task, the more such steps. More precisely, if tasks could be broken down into a long sequence of equal-length subtasks with a constant (and independent) chance of failure, such that to succeed in the whole task, the agent needs to succeed in *all* subtasks, then that would create an exponential survival curve. I.e. when Pr(*Task*) = Pr(*Subtask$_1$* & *Substask$_2$* & … & *Subtask$_N$*).

But we don't have to assume a perfect breakdown of the task into equal-length-equal-difficulty subtasks in order to get an exponential distribution (and corresponding constant hazard rate). We can also think of the constant hazard rate model as being agnostic as to how the task is broken down, just saying that the chance of succeeding in a subtask of duration *t* is always equal to the chance of succeeding in each of a set of smaller subtasks whose combined duration is also *t*. So on this model, it doesn't matter what level of granularity you assess the task at — whether you see it as one 60-minute task, six 10-minute tasks, or a 20-minute task plus forty 1-minute tasks — the chance of succeeding is set by the total time it would take a human to complete it.

This constant hazard rate model would predict that the time horizon for an 80% success rate is about ⅓ of the time horizon for a 50% success rate. This is because the chance of surviving three periods with an 80% success rate = $(0.8)^3$ = 0.512 ≈ 50%. More precisely, the time horizon for a success probability of *p* would be ln(*p*)/ln(*q*) times as long as one with success probability *q*. So an 80% time-horizon would be ln(0.8)/ln(0.5) = 0.322 times as long as the 50% time-horizon. Kwa et al. estimate the 80% time-horizon for Claude 3.7 Sonnet to be 0.25 as long, which is close to this theoretical estimate and within the margin of error given the noisiness of the results. The following chart shows how these numbers relate to the exponential survival curve (where $T_{50}$ is the 50% time-horizon etc.).

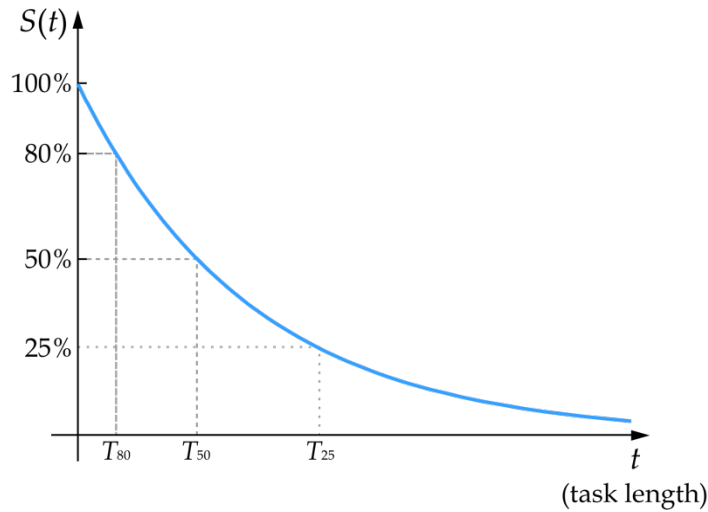

*Figure 3.* $T_{80}$, $T_{50}$, and $T_{25}$ represent the respective time horizons corresponding to an 80%, 50%, and 25% success rate.

Here are some useful comparisons for how the predicted time horizons over which an agent could get very high success rates compare to the measured time horizon for a 50% success rate:

$T_{80} \approx 1/3 \; T_{50}$
$T_{90} \approx 1/7 \; T_{50}$
$T_{99} \approx 1/70 \; T_{50}$
$T_{99.9} \approx 1/700 \; T_{50}$
*[each additional 'nine' of reliability beyond this divides the time horizon by 10]*

We can also use this model to calculate how long we'd expect it to take between the 50% success rate time horizon reaching a given length and a high success rate time horizon reaching that some length (on the assumption of the 7-month doubling time and the constant hazard rate model):

$T_{80}$ reaches any particular length about 1 year after $T_{50}$ does
$T_{90}$ reaches any particular length about 2 years after $T_{50}$ does
$T_{99}$ reaches any particular length about 4 years after $T_{50}$ does
$T_{99.9}$ reaches any particular length about 6 years after $T_{50}$ does
*[each additional 'nine' of reliability requires 2 more years…]*

Kwa et al. also attempt to fit a relationship between the success probability and time-horizon (see the figure below, which I've adapted from their paper). They plot the time horizon on a log scale and note that this reveals a sigmoid-shaped decay curve of success rate (the coloured bars). They show that this is reasonably well fit by a logistic function (the black curves). The paper doesn't compare this to how well alternative functions would fit the data.

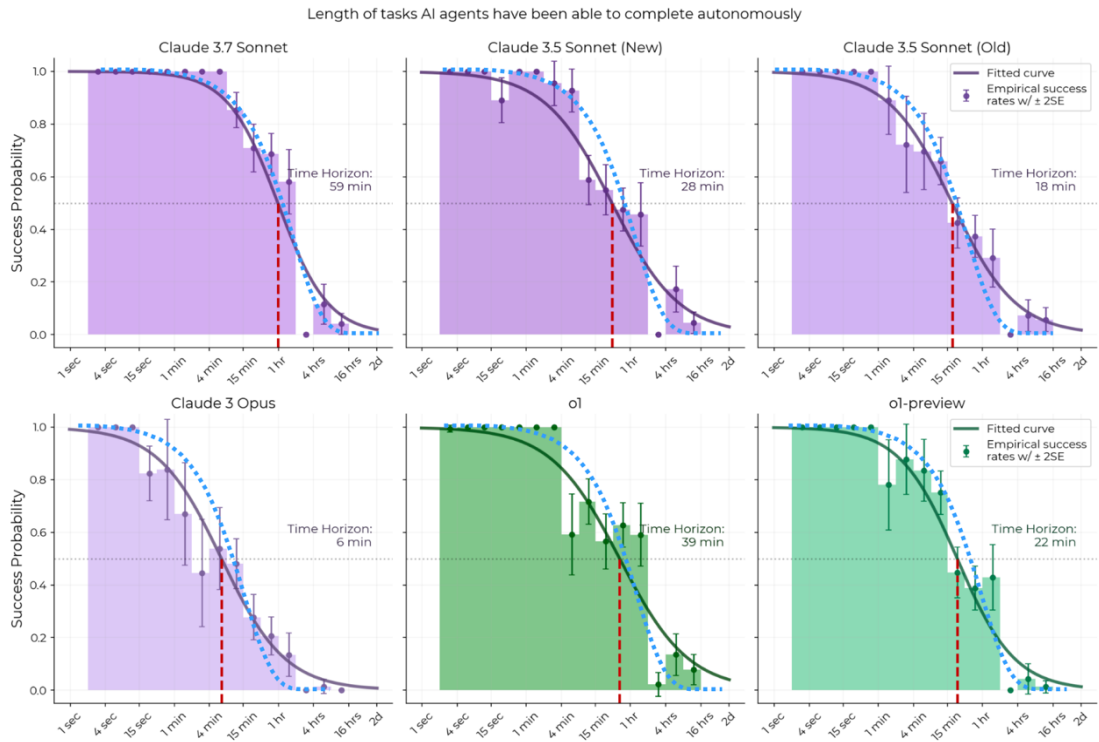

*Figure 4.* The decline in success rate with increasing task-duration for six agents. The coloured bars show the success rate of the model on tasks that took a certain duration for humans to solve, the black curve is the log-logistic distribution fitted to this data by Kwa et al, and the dotted blue curve is an exponential distribution that I have fit to the data. (Adapted from Kwa et al.)

The data is also well-fit by an exponential function, which also looks like a sigmoid when plotted on this logarithmic x-axis. I've added this to the above figure (from Kwa et al.) in the form of the dotted blue curves. It fits the data better for some models (such as the top left) and worse on others. It fits the data roughly as well overall, while being substantially more likely *a priori* — exponential decay is *the* simplest survival curve and has only one free parameter instead of two. Moreover, while logistic functions are quite simple and natural, the black curve here is not really a logistic distribution, but the more complex log-logistic distribution which merely looks like a logistic distribution when plotted on a logarithmic x-axis.

The paper also plots the survival curve for *human* performance over increasing time periods:

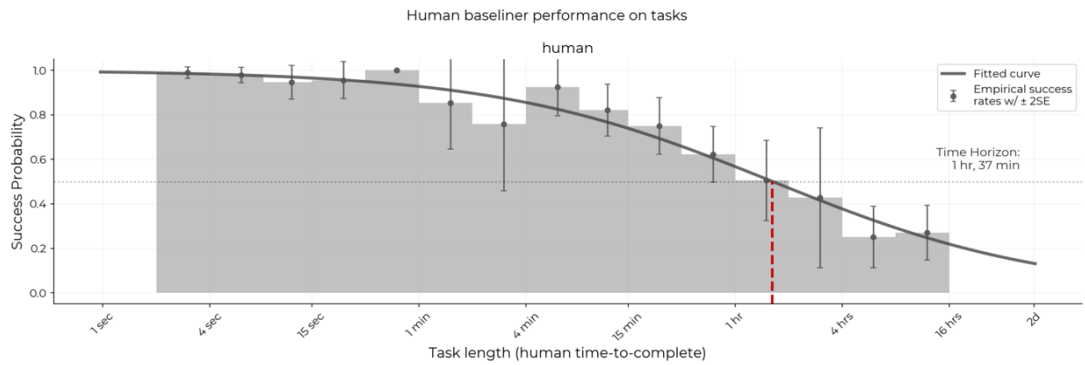

*Figure 5.* The decline in success-rate for humans as the task duration increases (from Kwa et al).

Intriguingly, this human survival curve seems to be noticeably better than a constant hazard rate (i.e. the chance of succeeding over long timescales drops off more slowly for humans than the constant hazard rate predicts). For example, on this graph, humans had about a 50% success rate at the 1.5-hour mark, which suggests 25% at 3 hours, 12.5% at 6 hours and 6.25% at 12 hours if the hazard rate is constant. However, the humans were still above 20% success rate at that point.

This could indicate a different scaling behaviour of success rate with time horizon for humans compared to AI agents, which would be well worth investigating and may suggest important underlying mechanisms (e.g. that the humans were better at correcting earlier failed subtasks). If human performance scales differently with task length than AI agent performance, that would be an important result, suggesting that there is a notable inefficiency in the current AI paradigm. This warrants further research.

However, there are other potential explanations. For instance, it could also be an artefact of this graph being an aggregate of humans with different ability levels, since even if all individual humans have a constant hazard rate (and so each have an exponential survival curve) a mixture of different humans would be a weighted sum of exponentials with different time constants and that distribution decays slower than an exponential (see Ord (2023) for details).[1]

---

[1] Because there is a separate curve estimated for each model, they don't suffer from mixing different capability levels into the same statistics. However, there is a similar effect that could still be present in the measurement of AI agents' performance over increasingly long tasks. It is plausible that some tasks are inherently easier than others per unit time, corresponding to different hazard rates. If so, then the survival curve over the whole task suite would be averaging different exponential decay curves together. This produces an aggregate decay curve with thicker tails than an exponential (corresponding to a declining effective hazard rate). Again, see Ord (2023) for details. Dealing with this effect is important as it could still be the case that the mechanism of constant hazard rate (and what it implies about the agents' behaviour) holds for every task, even if it isn't visible in the aggregate of these tasks.

**Upshots of the constant hazard rate model**

If the constant hazard rate model is sufficient to explain the drop-off in success rates on a task suite, there are several interesting upshots:

- It allows us to make predictions for the time-horizons at other success rates, such as 90% and 99%.

- It allows us to make predictions for how the success rate improves over time for a fixed task-length.

- It provides simple rules of thumb for predicting success probabilities (e.g. that if you double the task duration, you square the success probability).

- It suggests that AI agent performance (at least on this task suite) can be characterised by a half-life.

- It provides indirect evidence that what really is going on under the hood is that tasks are made up of many sequential subtasks and the chance of succeeding at the whole requires succeeding at every individual component. Moreover, this suggests that the current AI agents are not very good at recovering from earlier mistakes.

- Because the exponential distribution is the unique memoryless distribution, another way of seeing it is that the chance of failing at the next moment is independent of how far you've come — just like how the chance of a radioisotope decaying in the next minute is independent on how many minutes it has survived so far. This would be a surprising and interesting property for reasoning agents.

- Deviations from exponential decay for certain models may provide evidence that their hazard rate is increasing (or decreasing) with time, which might provide hints as to what they are doing wrong (or right).

- It can explain the same 7-month doubling time for the 50% time-horizon and 80% time-horizon: a 7-month halving-time for the underlying hazard rate would produce the 7-month doubling-time of all such time horizons.

- It also helps conceptually explain what the time horizons could even mean. For example, if it can complete a day's work, why can't it just do that twice to produce two days of work? On the constant hazard rate model, the issue is that if has a 50% chance of succeeding on Monday's work, then it only has a 25% chance of succeeding in both Monday's and Tuesday's work, which is too low to count as reliably achieving the 2-day task (and similarly for any higher reliability threshold).

Note that I am not claiming AI agents have a precisely constant rate of failure per minute of time it would take a human to complete the task. The claim is instead that something like this appears to be roughly true or stochastically true. All other models imply that there is some systematic change in the hazard rate over time, and my suggestion (pending more information about the precise fits of different models) is that the data doesn't warrant such assumptions.

If systematic deviations from exponential decay are found, such as the hazard rate increasing (or decreasing) with time, this might provide useful hints as to what the agents are doing wrong (or right). i.e. the constant hazard rate model can also work as a theoretical baseline from which to measure empirical deviations.

**Further work**

So far these results and analysis are merely suggestive of a constant hazard rate. Ideally one would conduct a formal statistical analysis on how well an exponential decay curve fits METRs data compared to the log-logistic they use. It would also be good to run robust statistical comparisons of the human decay curves versus the AI agent decay curves to see if there are systematic differences (e.g. different half-lives or different shapes that aren't just an artefact of the experimental setup). And of course it is also important to know how much any of this generalises to other suites of tasks.

**References**

Thomas Kwa et al., [Measuring AI Ability to Complete Long Tasks](), arXiv:2503.14499 [cs.AI], 2025.

Toby Ord, [The Lindy Effect](). arXiv:2308.09045 [physics.soc-ph], 2023.